\documentclass{article}
\usepackage{spconf,amsmath,graphicx}

\pdfoutput=1

\usepackage{amsfonts,amssymb}
\usepackage{algorithm}
\usepackage[noend]{algpseudocode}
\usepackage{caption}
\usepackage{listings}
\usepackage[utf8]{inputenc}
\usepackage{color}
\usepackage{xspace}
\usepackage{bbm}
\usepackage{dsfont}
\usepackage{mathtools}
\usepackage{amssymb}

\usepackage[table]{xcolor}
\usepackage{xspace}
\usepackage{tabularx}
\usepackage[justification=centering]{subfig}
\usepackage{url}
\usepackage{balance}
\usepackage{booktabs}
\usepackage[inline]{enumitem}

\def\iff{\xLeftrightarrow}
\graphicspath{{figs/}}

\def\ct#1{{\Gamma_{#1}}}
\def\ctt#1{{\Psi_{#1}}}
\def\cto{{\ct{}^*}}
\def\dt#1{{\phi_{#1}}}
\def\dth#1{{\hat{\dt{#1}}}}

\renewcommand{\Re}[1]{\mbox{$\mathbbm{R}^{#1}$}}

\def\R2{\Re{2}}

\def\norm2#1{\|#1\|}
\def\nnorm#1{|#1|}

\def\dg{{d_g}}
\def\chull#1{{ch(#1)}}

\newcommand{\argmin}{\arg\min}

\newdimen\origiwspc
\newdimen\origiwstr
\origiwspc=\fontdimen2\font 
\origiwstr=\fontdimen3\font 

\def\eg{\emph{e.g.}}

\def\eg{\emph{e.g.\xspace}}

\def\etal{\emph{et al.\xspace}}
\def\ie{\emph{i.e.\xspace}}

\title{Geometric Median Shapes}
\name{Alexandre Cunha\thanks{Funding supporting this work was provided by the
Beckman Institute at Caltech, a research center endowed with funds from the
Arnold and Mabel Beckman Foundation, to the Center for Advanced Methods in
Biological Image Analsysis.  Correspondence should be addressed to
\texttt{cunha@caltech.edu}.}}

\address{\vspace{-0.05cm}
Center for Advanced Methods in Biological Image Analysis\\ 
Center for Data Driven Discovery \\California Institute of Technology, Pasadena, CA, USA}

\begin{document}
\ninept
\maketitle
\begin{abstract}

We present an algorithm to compute the geometric median of shapes which is
based on the extension of median to high dimensions. The median finding problem
is formulated as an optimization over distances and it is solved directly using
the watershed method as an optimizer. We show that the geometric median shape
faithfully represents the true central tendency of the data, contaminated or
not.  It is superior to the mean shape which can be negatively affected by the
presence of outliers.  Our approach can be applied to manifold and non manifold
shapes, with single or multiple connected components. The application of
distance transform and watershed algorithm, two well established constructs of
image processing, lead to an algorithm that can be quickly implemented to
generate fast solutions with linear storage requirement. We demonstrate our
methods in synthetic and natural shapes and compare median and mean results
under increasing outlier contamination.

\end{abstract}

\vspace{-0.2cm}
\section{Introduction}
\label{sec:intro}

The computation of an average shape from a collection of similar shapes is an
important tool when classifying, comparing, and searching for objects in a
database or in an image. It is helpful, for example, when morphology is key in
the identification of healthy and diseased cells, when differentiating wild
from mutated cells, in the construction of anatomical atlases, and when
combining different candidate segmentations for the same image. The average
usually takes the form of the mean or the median, the latter being preferred
due to its robustness to automatically filter out outliers. Shape analysis is a
vast field of research and many solutions have been proposed to determine
geometrical and statistical mean shapes (see \eg\ \cite{klassen2004,da2018}),
with applications ranging from cell biology to space exploration and methods in
the fields of elastic shape analysis \cite{mio2007,sri2011} and most recently
in deep learning for shape analysis \cite{boscaini2016,vestner2017,groueix2018}.

Comparatively, very few have proposed methods for computing medians of shapes.
Fletcher \etal\ \cite{fletcher2008} developed a method to compute the geometric
median on Riemannian shape manifolds showing robustness to outliers on the
Kendall shape space. In \cite{berkels2010} the authors formulate the median
finding problem as a variational problem taking into account the area of the
symmetric difference of shapes, a model investigated much earlier in
\cite{cunha2004shape}. More recently, the work in \cite{hu2018} represents
shapes as integral currents and consider median finding using a variational formulation
which the authors claim lead to an efficient linear program in practice. They
observed that smooth shapes do not necessarily lead to smooth medians, an
aspect we also verified in some of our experiments.

We propose a method to quickly build median shapes from closed planar contours
representing silhouettes of shapes discretized in images. Our median shape is
computed using the notion of geometric median \cite{haldane1948}, also known as
the spatial median, Fermat-Weber point, and $L_1$ median (see \cite{small1990}
for the history and survey of multidimensional medians), which is an extension
of the median of numbers to points in higher dimensional spaces. The geometric
median has the property, in any dimension, that it minimizes the sum of its
Euclidean distances to given anchor points $x_j\in\Re{n}$,
\begin{equation}
\label{eq:geomedian}
      x^* = \argmin_{x} \sum_j \norm2{x - x_j}\quad .
\end{equation}
It has a breakdown point of 0.5, \ie\ the data can be corrupted with up to
50\% and it still gives a good prediction representing the sample
\cite{lopuhaa1991} (and \cite{gervini2008} for the functional median).  Unlike
the mean (centroid) of points, which can be found directly using a analytical
expression, the geometric median can only be approximated by numerical
algorithms. Due to convexity of eq.\eqref{eq:geomedian} the median point always
exist and it is unique provided the anchor points are not collinear
\cite{kemperman1987}.  The geometric median lies within the convex hull of the
given points (see \eg\ \cite{minsker2015}), a property we will exploit in our
approach.

Researchers have developed fast numerical solutions for this convex problem.
Weiszfeld developed in 1937, at the age of 16 years old, a gradient descent
algorithm which is still quite competitive (see translated reprint
\cite{weiszfeld2009}, and \cite{beck2015} for a review). It was
rediscovered many times until recognition on early 60's. New developments
claim faster algorithms that can handle large data sets \cite{vardi2000,
cardot2013}. We do not rely on these methods as our median shape is generated
by finding paths of minimum cost in the discrete image domain.

The emphasis of our presentation is to introduce a novel algorithm
capable of generating median shapes in a fast manner using tools already
familiar to the image processing community. We will leave proofs for
the mathematical models and formulations presented here for future work.

%
\section{Methods}
\subsection{A distance for planar shapes}
\label{sec:distance}

We define here the expression to measure the distance between closed planar
shapes represented by their contours. It is this metric that we will be
optimizing when computing the geometric median of a set of shapes. We assume
contours are rectifiable curves, but do not restrict their topology to planar
manifolds, crossings are allowed.  Other popular metrics to measure distances
between shapes, \eg\ Hausdorff and Fr{\'e}chet distances, are not suitable for
our median formulation as they do not report distances along the entire length
of candidate curves.

\begin{figure}[t!]
\begin{minipage}[t]{\linewidth}
  \centering
  \includegraphics[width=\columnwidth]{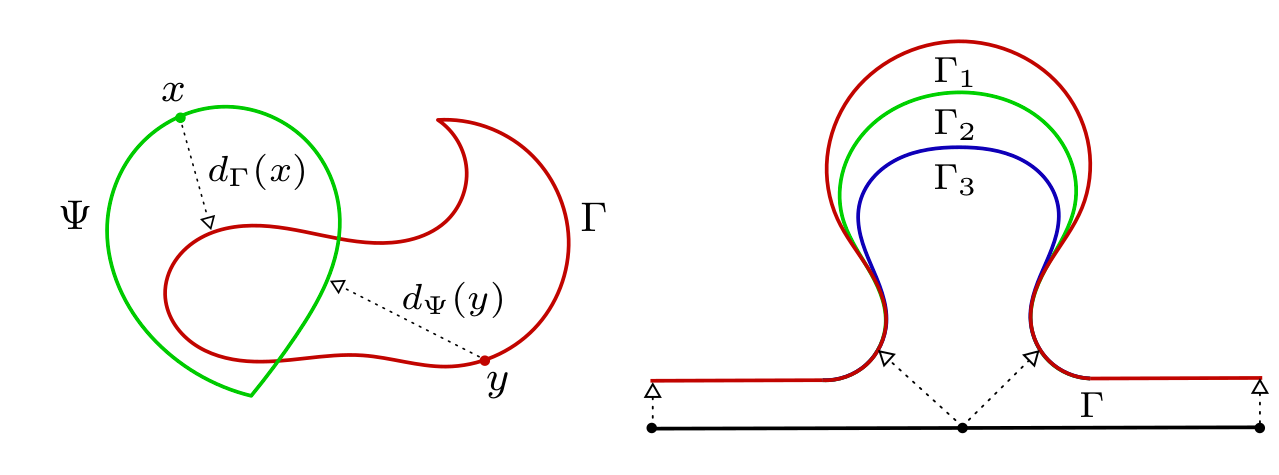}
\end{minipage}
\vspace{-0.20in}
\caption{\label{fig:distance}\footnotesize {\bf Illustration of planar shape
distances}. Arrows in the left figure represent the shortest distances $d_{\ct{}}(x)$ and
$d_{\Psi}(y)$ from points $x\in\Psi$ and $y\in\ct{}$, respectively, to closed contours
$\ct{}$ and $\Psi$.  The figure on the right illustrates a case where the
horizontal black curve $\ct{}$ is equally distant to three different other curves,
$d(\ct{},\ct{1}) = d(\ct{},\ct{2}) = d(\ct{},\ct{3}) = 63$. The symmetric
distance $\dg$ captures our two-way expectations: $\dg(\ct{},\Gamma_1) = 145 >
\dg(\ct{},\ct{2}) = 129 > \dg(\ct{},\ct{3})= 116$. Curves $\ct{i}$ only differ
at the bulgy part.  Slanted arrows on the right figure are directed to the
farthest points on $\ct{i}$ from $\ct{}$. To compute $\dg$, curves were
8-connected discretized in 429x295 images.} {\color{lightgray}\hrule}
\vspace{-0.2in}
\end{figure}

Let $d_{\ct{}}\colon\R2\to\Re{}$ give the distance of a point
$x\in\R2$ to a curve $\ct{}\subset\R2$,
\begin{equation}
\label{eq:distpoint}
  d_{\ct{}}(x) = \inf_{y\in\ct{}} \norm2{x -y} \geqslant 0
\end{equation}
where the infimum is attained for one or more points $y\in\ct{}$ closest to
$x$, and the $L_2$ norm $\norm2{\cdot}$ gives the Euclidean distance between
points (see Fig.\ref{fig:distance}). Note that $d_{\ct{}}(x) = 0\iff\ x\in\ct{}$.  We then define the
distance from a rectifiable curve $\ct{}$ to another curve
$\ctt{}$ using a line integral over all points $x\coloneqq x(s)\in\ct{}$,
\begin{equation}
  \label{eq:distance}
  d(\ct{},\ctt{}) = \frac{1}{|\ct{}|}\int_{\ct{}} d_{\ctt{}}{ \left(x\right)} ds
\end{equation}
where $d_{\ctt{}}(\cdot)$ is given by eq\eqref{eq:distpoint}, $ds$ is the arc
length differential, and $|\ct{}| > 0$ is the length of curve $\ct{}$.  By
integrating distances for all points along the curve and dividing by its
length we have a measure that tells us how far, in average, is the source curve $\ct{}$
to the target curve $\ctt{}$. This measure is independent of curve parameterization.
%
%
It follows that
$d(\ct{},\ctt{}) \geqslant 0$ and $d(\ct{},\ctt{}) = 0\iff\ \ct{} = \ctt{}$. If
$\ct{}$ and $\ctt{}$ are parallel curves for which $\forall x\in\ct{},
d_{\ctt{}}(x) = \delta$, then $d(\ct{},\ctt{}) = \delta$, which is what we
intuitively expect. As an example, for two concentric circles with {\it radii}
$r_1 > r_2$, we have $\delta = r_1 - r_2$.  Another property of
$d(\cdot,\cdot)$ is its invariance under isometries (rigid transformations
including reflections), owing to the intrinsic property of curve length and to
the preservation of point distances under such transformations.

In general, this distance is asymmetric, $d(\ct{},\ctt{}) \neq d(\ctt{},\ct{})$,  
unless the curves themselves are symmetric with respect to an 
axis.  Similar to the Hausdorff distance, we combine $d(\ct{},\ctt{})$ and  
$d(\ctt{},\ct{})$ to obtain a symmetric distance $\dg$,  
\begin{equation}
  \dg(\ct{},\ctt{}) = \frac{1}{2}\left( d(\ct{},\ctt{}) + d(\ctt{},\ct{}) \right)
\end{equation}
possessing the following properties:
\begin{enumerate*}[label=\;(\arabic*)\;]
    \item $\dg(\ct{},\ctt{}) \geqslant 0$ (non--negativity);
    \item $\dg(\ct{},\ctt{}) = 0\iff\ \ct{} = \ctt{}$ (identity);
    \item $\dg(\ct{},\ctt{}) = \\ \dg(\ctt{},\ct{})$ (symmetry);
    \item $\dg(\ct{},\ctt{}) \leqslant \dg(\ct{},\Phi) +
      \dg(\Phi,\ctt{}),\;\forall \ct{},\\ \ctt{},\Phi$ (triangle inequality).
\end{enumerate*}
Properties (1) to (3) are due to the Euclidean norm used in $d(\cdot,\cdot)$.
The triangle inequality property requires formal proof, a task for future
work.  All properties have been experimentally validated.

\subsection{Geometric median for shapes}
\label{sec:gmedian}

We formulate the problem of finding the geometric median of shapes as an
optimization problem over distances, similar to the high dimensional geometric median of points:
find the optimal closed contour $\cto$ that minimizes the sum of
distances to a given set of closed contours $\{\ct{i}\}_{i=1}^{n}$,
\begin{equation}
\label{eq:opt}
  \cto = \argmin_{\ct{}} \sum_{i=1}^{n} d(\ct{},\ct{i})
\end{equation}
where $d(\cdot,\cdot)$ is given by eq.\eqref{eq:distance} and the solution
lies in the convex hull of $\{\ct{i}\}$, $\cto\subset\chull{\{\ct{i}\}}$.  When
each curve $\ct{i}$ degenerates to a point, with $\nnorm{\ct{i}} = 1$, then the
problem is reduced to finding the geometric median of points. From
eq.{\eqref{eq:opt} we then have
\begin{equation}
\label{eq:star}
\begin{split}
  \cto & = \argmin_{\ct{}}\sum_{i=1}^{n} \left( \frac{1}{|\ct{}|}\int_{\ct{}} d_{\ct{i}}{ \left(x\right)} ds \right),\; x\in\ct{} \\
  & = \argmin_{\ct{}} \frac{1}{|\ct{}|}  \int_{\ct{}} \left( \sum_{i=1}^{n} d_{\ct{i}}{ \left(x\right)}\right) ds \\
  & = \argmin_{\ct{}} \frac{1}{|\ct{}|}  \int_{\ct{}} \left( \sum_{i=1}^{n} \norm2{x - y_i} \right) ds,\; y_i\in\ct{i}
\end{split}
\end{equation}
where $\int$ and $\sum$ commute due to Tonelli's theorem for
non-negative functions \cite{folland2013} (indeed, $d_{\ct{i}}(\cdot)\geqslant 0$) and
$y_i$ is the point in $\ct{i}$ closest to $x\in\ct{}$, as in
Fig.\ref{fig:distance}. It should be clear that the
location of each $y_i$ depends on the position of $x$ in $\ct{}$, thus
$y\coloneqq y_i(x)$.

Let's for a moment ignore
$\nnorm{\ct{}}$ on eq.\eqref{eq:star} and consider an approximation of the
unknown $\ct{}$ by an ordered set of points
$(x_1,\ldots,x_m)$. At the solution $\cto \coloneqq (x_1^{*},\ldots,x_m^{*})$, the
term in parentheses on the last expression in eq.\eqref{eq:star} is optimized for each $x_j$ which implies
that any point in $\cto$ is in fact the geometric median of points $\{y_i\}_{i=1}^n$,
each from a single $\ct{i}$,
\begin{equation}
  x^* = \argmin_{x} \sum_{i=1}^{n}\norm2{x - y_i},\qquad y_i\in\ct{i}, \forall x^*\in\ct{}^*
\end{equation}
Since we know $x^*\in\chull{\{y_i\}} \subset \chull{\{\ct{i}\}}$, see \eg\
\cite{minsker2015}, we can then show by contradiction that
$\cto\subset\chull{\{\ct{i}\}}$, as stated above. This is a property that will
help us in the optimization. Directly computing each $x^*$ from points $y_i(x)$
is not straightforward due to the nonlinear dependency of $y_i$ on $x$. It is
also not clear how to properly select a set of $n$ points $\{y_i\}$, one for each
curve $\ct{i}$, to produce each median point $x_j^*$ in $\cto$.

By discarding the contribution of the curve length $\nnorm{\ct{}}$ and
optimizing solely based on distances, \ie\ minimizing the following functional
\begin{equation}
\label{eq:jcost}
J(\ct{}) = \int_{\ct{}} \left( \sum_{i=1}^{n} \norm2{x - y_i(x)} \right) ds, \quad x\in\ct{}
\end{equation}
we end up with a simpler problem while, surprisingly, still solving the
original optimization problem in eq.\eqref{eq:opt}. Our results consistently
demonstrate we are also optimizing for the curve length even though we are not explicitly
considering it in the optimization.

When we use a squared norm in eq.\eqref{eq:jcost} and elsewhere, \ie\ squared distances
$\norm2{x - y_i}^2$ instead of $\norm2{x - y_i}$, then
we call $\cto$ the optimal mean shape. We compare mean and median shapes in our
experiments and show the latter is much less sensitive to outliers.

\subsection{Optimization with watershed transform}
\label{sec:opt}

We look for the closed contour having overall the smallest sum of distances
from its points to all given contours $\ct{i}$.  We solve this problem as an
optimal path finding problem where the watershed with markers method
\cite{beucher1992} is applied over the accumulated distance transforms image of the
given curves to generate the median contour $\cto$.  Briefly, the
basic idea to form the median is as follows. For every grid point $x$ in the
image domain we find its total distance $\dt{}(x)$ to all the contours $\ct{i}$
-- this is what the accumulated distance transform does. The next step is to
find those grid points forming a closed path which together have the least sum
of total distances, $\dt{}(\cto) < \dt{}(\ct{}), \forall\ct{}\neq\cto$.  This
is accomplished by the watershed method which computes the closed path with
minimum cost.

Others have applied the watershed method
to solve optimization problems on either a graph or a discrete grid. For
example, Cuprie \etal\ \cite{couprie2011surface} used the Power Watershed
\cite{couprie2011power} in their formulation of the surface reconstruction from
point cloud problem to minimize a weighted total variation functional where
weights are proportional to Euclidean distances between points.

We abuse the notation here and use $\ct{i}$ to designate both the 8-connected
contour and the image containing it.  We represent each contour $\ct{i}$ as the
zero level set of a unsigned distance transform function
$\dt{i}\colon\Omega\to\Re{}_{+}$, such that $\dt{i}(x) = 0$ for $x\in\ct{i}$,
$\dt{i}(x) > 0$ for $x\in\Omega\setminus\ct{i}$, and $\dt{i}(x)$ gives the
shortest distance from $x$ to curve $\ct{i}$. Thus, from
eq.\eqref{eq:distpoint}, $\dt{i}(x) = d_{\ct{i}}(x)$. If we take the sum
$\dt{}(x) = \sum_{i=1}^{n} \dt{i}(x)$ for every point $x$ in the domain
$\Omega$ of the image, {\em the median contour $\cto$ is a minimal path in the image
$\dt{}$}.

To create markers for the watershed, we take the union of the basins in
$\dth{}$ (the inverted $\dt{}$) with the complement of the convex hull of
contours $\ct{i}$, $\Omega\setminus\chull{\{\ct{i}\}}$. The marker imposed
outside the convex hull prevents the creation of false positive regions due to
eventual basins placed close to the boundary of the image.  Points in regional minima
are those farthest from the contours, and hence serve well as markers.  The
watershed lines computed from these markers form $\cto$, as presented in the GEMS
algorithm below.
{\noindent
\begin{minipage}{\columnwidth}\vspace{-0mm}
\begin{algorithm}[H]
\renewcommand{\thealgorithm}{}
\floatname{algorithm}{}
\caption{\bf GEMS -- Geometric Median Shapes}\label{alg:gems}
\begin{algorithmic}[1]
\Procedure{GEMS}{$h, \ct{1},\ldots,\ct{n}$}
\State{\textbf{Input}: $h$ (minima height), $\ct{i}$ (contour images)}
\State{\textbf{Output}: $\cto$ (geometric median)}
\State{$\dt{} = 0$}
\For{$i = 1,\ldots,n$}
\State{$\dt{}' = \text{EuclideanDistanceTransform}(\ct{i})$}
\State{$\dt{} = \dt{} + \dt{}'$}
\EndFor
\State{{\bf end}}
\State{$\dth{} = \text{invert}(\dt{})$}
\State{$B = \text{WatershedBasins}(\dth{}, h)$}
\State{$H = \text{ConvexHull}(\ct{1},\ldots,\ct{n})$}
\State{$markers = B\cup\{\Omega\setminus H\}$}
\State{$\cto = \text{Watershed}(\dth{}, markers)$}
\EndProcedure
\end{algorithmic}
\end{algorithm}
\end{minipage}
\vspace{-3mm}
\begingroup
\captionof*{algorithm}{\footnotesize Our current implementation of GEMS is a
script that uses distance transforms, height minima, minima, convex hull, and
watershed implementations from the {\tt pink} \cite{pink} and  {\tt gmic}
\cite{gmic} packages, with image
I/O operations primarily supported by {\tt gmic}. Note that
memory space is linear in image size and does not grow with the number of
contours $\ct{i}$. The only parameter that needs adjustments is the height $h$
used for the detection of watershed basins. The {\tt\bf{for}} loop can be 
parallelized to speed up computations.}
\vspace{-2mm}
{\noindent\color{lightgray}\rule{1.0\columnwidth}{0.2mm}}
\vspace{-1mm}
\endgroup

When the contours $\ct{i}$ are highly dissimilar in shape and location then a
shape representing the central tendency of the data is either too weak or non
existent.  This is reflected in the value of the height parameter $h$: the
smaller the $h$ value needed to create $\cto$, the less cohesive is the data --
provided colocalized shapes are not too slim. For example, we could not obtain a median
for the colored outliers in Fig.\ref{fig:circles}, even after taking $h = 1$,
as they are very dissimilar to each other and dispersed.

When contours reach the boundaries of the image, GEMS results need to be
augmented. This is, for example, the case when building a consensus out of many
binary contour segmentations $\{\ct{i}\}$ for tiles of an image -- see
Fig.\ref{fig:consensus}. Disconnected basins are formed at the boundary of the
image $\hat{\dt{}}$ and we no longer have the larger marker outside the convex
hull to eliminate them. This in turn leads to the creation of extra regions not
representing the actual data.  We repair this over segmentation, whenever
present, by removing edges from $\cto$ according to the following
criteria:
\begin{enumerate*}[label=(\arabic*)\;]
\item Let $\ct{}_{or} = \bigcup_{i}\ct{i}$. Edges in $\cto$ not present in $\ct{}_{or}$ are removed: if the edge
has no candidate from the pool of segmentations, there is no justification to keep
it; this eliminates the great majority of extraneous edges in $\cto$.
\item Edges in $\cto$ which have a small count (currently 20\%) in the pool
$\{\ct{i}\}$ and an average intensity close to background are 
eliminated; this is done to remove edges introduced only by a few users that can be potentially 
annotation mistakes.
\end{enumerate*}
\begin{figure}[!b]
\hrule
\begin{minipage}[t]{1.0\columnwidth}
  \centering
    \includegraphics[width=\columnwidth]{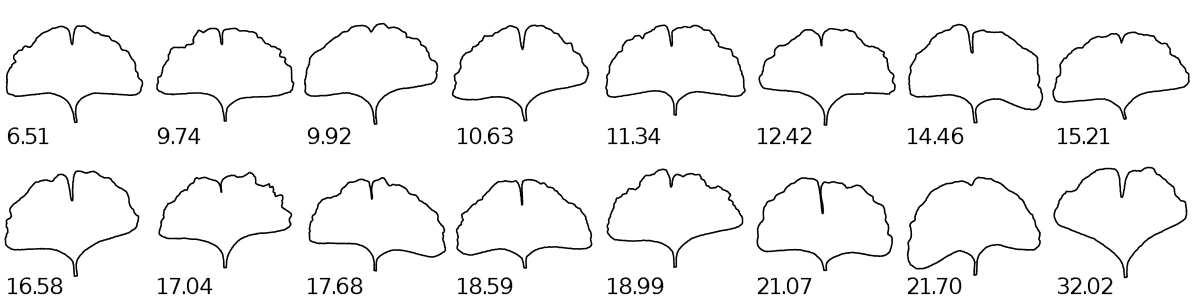}\\[1mm]
     \hrule
    \includegraphics[width=\columnwidth]{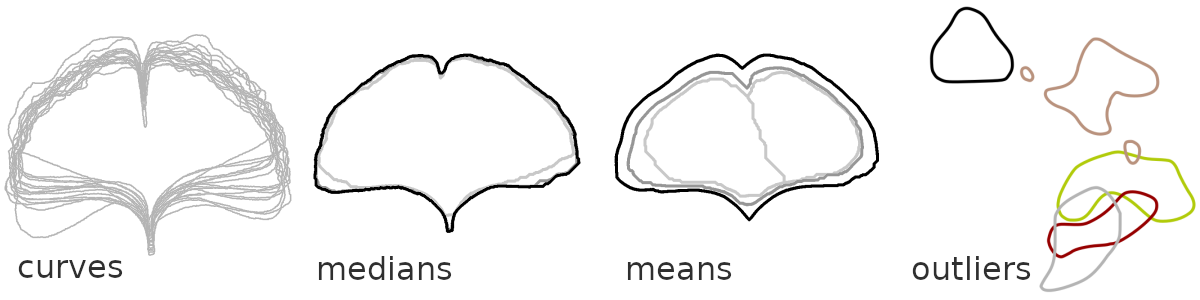}
\end{minipage}
\hrule
\vspace{2mm}
\caption{\label{fig:ginkgo}\footnotesize {\bf Median and mean shapes for {\emph{\textbf{Ginkgo
    biloba}}} leaves}. Sixteen leaf curves are ordered according to their $d_g$
  distances to their geometric median. They were midvein aligned and scaled
  keeping their original aspect ratio and shape, all shown on the {\em curves}
  panel. Their respective geometric median and mean shapes are the black curves
  on the {\tt medians} and {\tt means} panels. Median and mean are $2.63$ pixel
  units $d_g$ distant. Note that the {\it petiole} (lower tip) is better
  captured by the median. The gray mean shapes were obtained after tainting
  data with two (black and brown curves) and five outliers, shown on the
  rightmost {\tt outliers} panel, resulting in poor average shapes distant
  $24.11$ and $56.75$ from the true mean. We can no longer obtain a simple mean
  curve when these five outliers are considered. These outliers barely
  affect the median. The gray median, shown on the {\tt medians} panel, is only
  $10.38$ pixel units apart from the true median. It was obtained after
  contaminating data with ten outliers, twice as much used for the mean, thus
  illustrating the superiority of the geometric median. Leaf images have 800x800 pixels.}
\vspace{-4.5mm}
\end{figure}

Our experiments have consistently demonstrated that the median shapes $\cto$
obtained using GEMS have a cost $\sum_{i} d(\ct{},\ct{i})$ lower than any other
contour in its vicinity, $\ct{} = \cto + \epsilon\ctt{}, \epsilon\ll 1$, and
elsewhere. One might argue that the cost could be reduced, according to
eq.\eqref{eq:distance}, by picking a longer curve. But this is not the case.
Choosing a contour $\ct{}$ longer than $\cto$, a few pixels away from the
optimal configuration, increases the sum of distances $\dt{}$ at a greater
rate, as the accumulated point distances $\dt{}(\ct{})$ grows faster than the
curve length $|\ct{}|$.  Conversely, choosing a curve with fewer points to
decrease $\dt{}(\ct{})$ does not reduce the cost $J(\ct{})$ as now the curve
length is smaller and although we have now fewer points each has a higher
accumulated distance.  Evidences suggest then that GEMS not only finds the
contour with lowest accumulated $L_1$ distances, $J(\ct{})$, but it also picks
the path with the right length. This indicates that $\cto$ optimizes distances
$d(\cdot,\cdot)$, and consequently $\dg(\cdot,\cdot)$. We report values for
$\dg$ in our experiments.

\vspace{-2em}
\section{Experimental Results}
\label{sec:results}
We demonstrate GEMS on a set of planar shapes ranging from plant leaves to hand
traced circles and non-manifold shapes resulting from interactive segmentation of
biological cells. The examples in the figures are self contained and address
particular cases of interest. Leaf data is from \cite{mallah2013}.

In the {\it Ginkgo biloba} leaves experiment, Fig.\ref{fig:ginkgo}, we compute
median and mean shapes for sixteen aligned contours, with and without outliers.
Median and mean are initially pretty close, $\dg{} = 2.63$, but their distances
increase with the introduction of outliers. The influence of data alignment for
the median computation is presented in Fig.\ref{fig:tulip}, where tulip leaves
are registered in different ways leading to distinct but close in shape
medians. We illustrate in the circles example, Fig.\ref{fig:circles}, a case
that the geometric median holds the central tendency even when the
contamination with outliers is beyond 50\%. This is not the case for the mean
shape which deteriorates quickly as we increase the number of random outliers
tainting the data. In the last experiment, Fig.\ref{fig:consensus}, we use the
augmented GEMS described in section \ref{sec:opt}, to create a consensus
segmentation for a small set of interactively generated segmentation results.
These segmentations are for small tiles cropped from large cell images. We
compare our results with those created by the Staple method \cite{warfield2004}
and show that our approach is comparable, and sometimes better, and can produce
results much faster.
%
%
\begin{figure}[b!]
\vspace{-1em}
  \hrule
  \vspace{1mm}
  %
  \begin{minipage}[t]{\linewidth}
   \centering
   \includegraphics[width=\columnwidth]{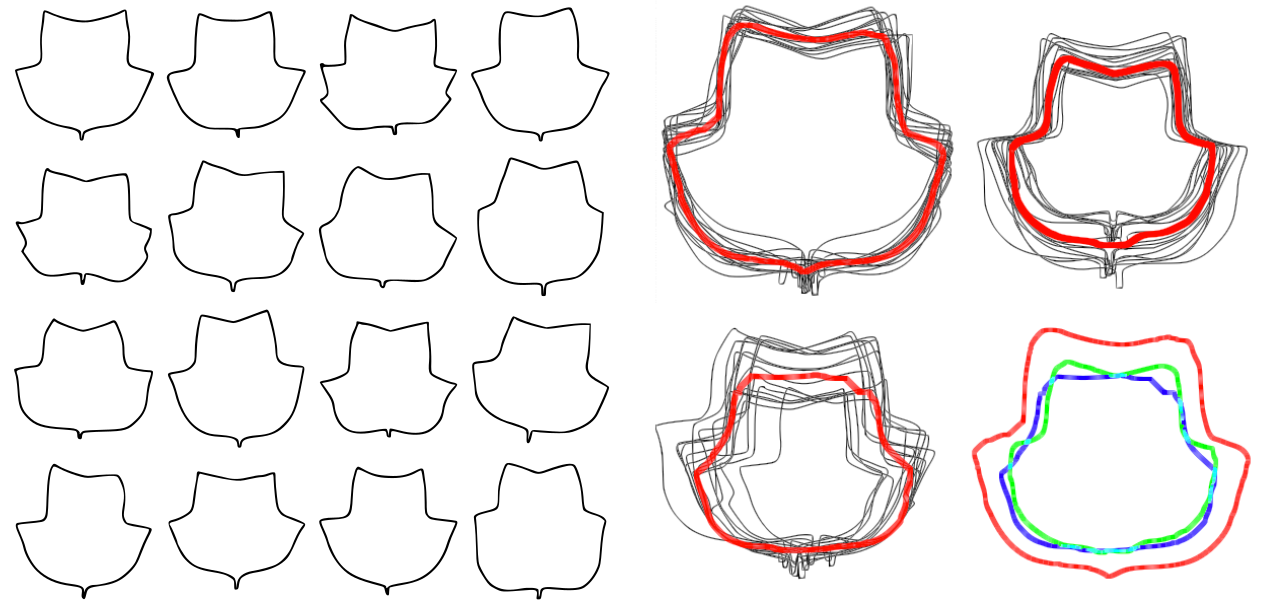}
  \end{minipage}
  \vspace{-2mm}
  \hrule
  \vspace{1mm}
  \caption{\label{fig:tulip}\footnotesize {\bf Alignment and
  medians}.  Sixteen tulip tree ({\it Liriodendron tulipifera}) leaves, shown
  on the left panel, are aligned in different ways leading to different median
  shapes (bottom right curves) conforming to the alignment. We overlay the
  medians, shown as thick red contours, onto the aligned leaf curves to
  visually demonstrate their placement: in the convex hull of the given shapes
  and not biased towards the extremes -- see \eg\ the top right
  case. Registrations were: no registration (top,left), registration to a line
  passing through side lobe corners (top right), and rigid registration, all
  done using the same chosen target image from from the pool.}
  \vspace{-4mm}
\end{figure}
\begin{figure}[t!]
\begin{minipage}[t]{\columnwidth}
    \includegraphics[width=\columnwidth]{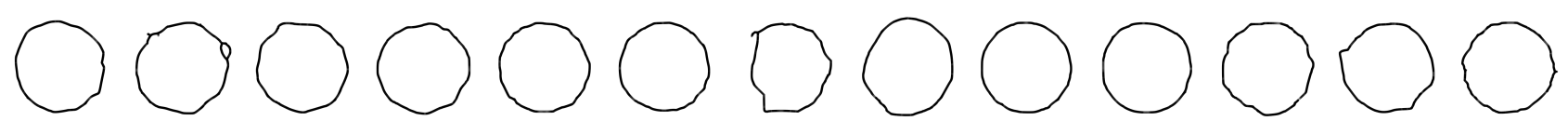}
    \includegraphics[width=\columnwidth]{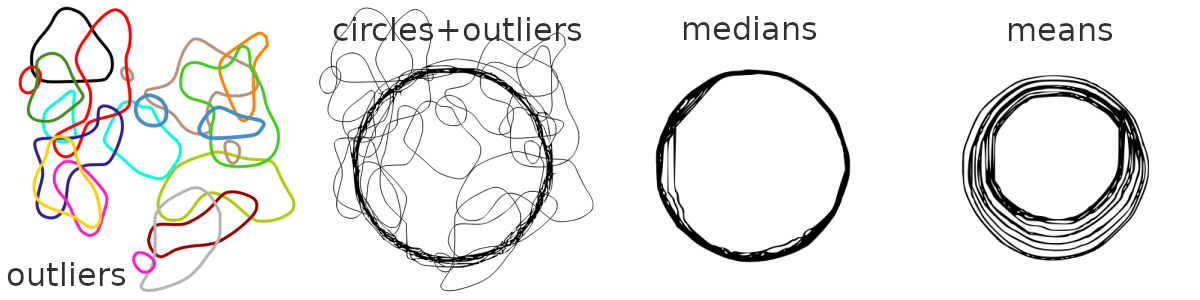}
\end{minipage}
\hrule
\vspace{1mm}
\begin{minipage}[t]{\columnwidth}\footnotesize
  %
  \begin{tabular*}{\columnwidth}{rrrrrrrr} 
  {\bf \#outliers} & 2 & 4 & 6 & 8 & 10 & 12 & 14 \\
  {\bf median} & 0.52 & 0.88 & 1.15 & 1.48 & 1.88 & 2.47 & 3.04 \\
  {\bf mean} & 9.91 & 20.47 & 27.72 & 35.13 & 40.01 & 43.70 & 47.69 \\
  \end{tabular*}
  %
  \hrule
\end{minipage}
\vspace{-3mm}
\caption{\label{fig:circles}\footnotesize \hyphenpenalty=10000 \exhyphenpenalty=10000
{\bf Robustness to outliers.} Thirteen noisy circles hand traced by different
users, shown on top row, were collected using our {\it Collaborative
Segmentation} web application.  We progressively added randomly generated
outliers to contaminate the circle data with up to fourteen outliers, {50\%+1}
contamination (colored {\em rubber bands} on {\tt outliers} panel, one color
per outlier, some disconnected). These are strong outliers as they greatly
differ in size, shape, topology, and location when compared to the circles --
see {\tt circles+outliers} panel. For each additional outlier added to the pool
of circles, we compute new median and mean shapes and then measure their
distance $\dg$ to, respectively, the median and mean shapes obtained without
outliers. These numbers are shown on the table above for seven out of fourteen
contaminations. The geometric median is marginally affected -- $\dg$ is at most 3 pixel
units -- while the mean significantly deteriorates, early and quickly. Panels
{\tt medians} and {\tt means} show respectively geometric median and mean shapes as
outliers progress, from one to fourteen. All processed images have 400x400 pixels. }
{\color{lightgray}\hrule}
\vspace{-0.25in}
\end{figure}
%
%
%
%
\begin{figure}[b!]
\begin{minipage}[t]{\columnwidth}\vspace{-4mm}
    \includegraphics[width=\columnwidth]{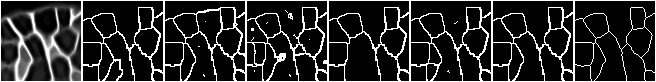}
    \includegraphics[width=\columnwidth]{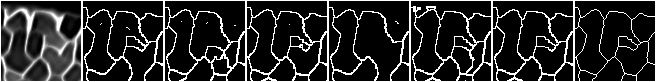}
    \includegraphics[width=\columnwidth]{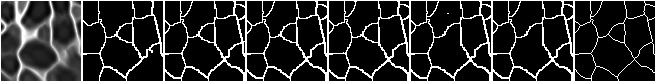}
    \includegraphics[width=\columnwidth]{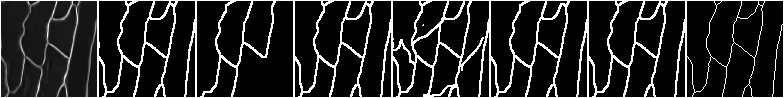}
\end{minipage}
\vspace{-3mm}
\caption{\label{fig:consensus}\footnotesize \hyphenpenalty=10000
  \exhyphenpenalty=10000 {\bf Segmentation consensus}. Results of interactive
  segmentation (columns 2 to 6) are shown for four small tiles (column 1) of
  much larger images. Although users create different segmentations for the
  same image, they nevertheless exhibit some agreement most prominent in clear,
  unambiguous parts of the image. Using our median approach to combine users'
  segmentations into a single one eliminates outliers and spurious annotations,
  \eg\ dangling scribbles and speckles, as shown in our results on the
  last column, 8.  Our solution is on par or better (third row) than results
  obtained with the Staple method \cite{warfield2004}, shown on column 7,
  and orders of magnitude faster (\eg\ 1.06s against 237.35s for
  sixteen 800x800 images) when compared to the expectation maximization
  approach adopted by Staple.  Users' segmentations are from our {\em
  Collaborative Segmentation} web application. $h = 5\ \text{or}\ 10$.}
\end{figure}

\vspace{-1em}
\section{Conclusions}
\label{sec:conclusion}
\vspace{-0.25em}
We presented an algorithm, GEMS, to compute geometric median shapes
which is simple, fast, straightforward to implement and requires adjustments of
a single parameter. The approach leads to median shapes that are robust to
outliers supporting a contamination of over 50\% while still capturing the
central tendency of given shapes. We adopted the watershed transform as the
optimizer to identify paths of minimal cost in the accumulated distance
transform image. Our experiments have shown that the proposed symmetric
distance $\dg$ is optimized for the geometric median contour even though the
curve length is not explicitly considered in the optimization. From our initial
experimentation, our approach compares well with the Staple method in the
segmentation consensus problem while being faster by many orders of magnitude.

As part of future work, we intend to extend the model and experiments
to 3D. We also plan to investigate
using distances other than the Euclidean distance which might lead to other
types of average shapes. We also plan to adopt weights to combine shapes, in a
weighted average fashion, as this might be useful in shape
clustering problems. We intend to investigate if and how the distance $\dg$ and
our median shapes can improve classification in large
data sets. We are also interested in continuing studying the potential of medians
for the segmentation consensus problem in both 2D and 3D, as the need to
combine individual segmentations while automatically discarding outliers is of
importance to real time collaborative segmentation systems.
%

%
%
%
\bibliography{main}
\end{document}